\documentclass[conference]{IEEEtran}
\IEEEoverridecommandlockouts
\usepackage[backend=biber]{biblatex}
\usepackage{amsmath,amssymb,amsfonts}
\usepackage{algorithmic}
\usepackage{graphicx}
\usepackage{textcomp}
\usepackage{xcolor}
\usepackage{hyperref}
\def\BibTeX{{\rm B\kern-.05em{\sc i\kern-.025em b}\kern-.08em
    T\kern-.1667em\lower.7ex\hbox{E}\kern-.125emX}}
\begin{document}

\title{A new face swap method for image and video domains: a technical report}

\author{\IEEEauthorblockN{Daniil Chesakov\text{*} \thanks{\text{*} Former employee}}
\IEEEauthorblockA{\textit{Sber AI} \\
Moscow, Russia \\
dchesakov@nes.ru}\\
\IEEEauthorblockN{Andrey Kuznetsov}
\IEEEauthorblockA{\textit{Sber AI, Samara National Research University} \\
Moscow, Russia \\
AVladimirKuznetsov@sberbank.ru}
\and
\IEEEauthorblockN{Anastasia Maltseva}
\IEEEauthorblockA{\textit{Sber AI} \\
Moscow, Russia \\
AAMittseva@sberbank.ru}\\
\IEEEauthorblockN{Denis Dimitrov}
\IEEEauthorblockA{\textit{Sber AI, MSU} \\
Moscow, Russia \\
Dimitrov.D.V@sberbank.ru}
\and
\IEEEauthorblockN{Alexander Groshev}
\IEEEauthorblockA{\textit{Sber AI} \\
Moscow, Russia \\
AYGroshev@sberbank.ru}
}
\maketitle

\begin{abstract}
Deep fake technology became a hot field of research in the last few years. Researchers investigate sophisticated Generative Adversarial Networks (GAN), autoencoders, and other approaches to establish precise and robust algorithms for face swapping. Achieved results show that the deep fake unsupervised synthesis task has problems in terms of the visual quality of generated data. These problems usually lead to high fake detection accuracy when an expert analyzes them. The first problem is that existing image-to-image approaches do not consider video domain specificity and frame-by-frame processing leads to face jittering and other clearly visible distortions. Another problem is the generated data resolution, which is low for many existing methods due to high computational complexity. The third problem appears when the source face has larger proportions (like bigger cheeks), and after replacement it becomes visible on the face border. Our main goal was to develop such an approach that could solve these problems and outperform existing solutions on a number of clue metrics. We introduce a new face swap pipeline that is based on FaceShifter architecture and fixes the problems stated above. With a new eye loss function, super-resolution block, and Gaussian-based face mask generation leads to improvements in quality which is confirmed during evaluation.
\end{abstract}


\section{Introduction}
Nowadays a lot of visual content is used over the internet for different purposes. An immense number of graphical instruments to operate with visual data like images and videos is available free of charge. This leads to stating new tasks for people who process the data: quality enhancement, compression, restoration and coloring old photos, etc. We can see that the effect of artificial intelligence (AI) progress also does not lay only in the scientific research field, but goes out of this scope and makes it available for everyone to apply state of the art (SOTA) AI techniques for everyone over easy-to-use applications and social networks. In recent years we moved forward from an ordinary list of image processing tasks to such applications as background replacement \cite{MODNet} (e.g., in video conferences, photo editing tools, etc.), makeup style transfer \cite{Chang2018PairedCycleGANAS}, facial attributes correction \cite{FacialAttr2019}, hair style tranfser \cite{shen2020interpreting}, face/head swap (Snapchat) \cite{FaceSwapSurvey2020} and others. The last application became of a big interest for researchers for different purposes especially for cinema/clip making and entertainment needs. Face swap in general is a procedure of taking two visual data sources (source and target) and replacing the target face with the source face. By visual data sources, we mean images and videos, so different data combinations are used. The most frequently used combinations are the source and target images and source images and target video. One of the well-known examples of deep fake technology appeared in 2019 when a viral scene from Home Alone with Macaulay Culkin’s face swapped with Silvester Stallone was distributed overall social networks and this started the rise of deep fake popularity. In 2021 DeepTomCruise TikTok account attracted the attention of the audience with high-quality deep fake videos. Although these videos were perfectly synthesized \cite{vincent_2021} not only by using SOTA generative deep learning face swap algorithms, several postprocessing steps, and using an actor with a similar face, the result looks amazing. 

In this report, we would like to observe our new face swap approach that can be used for image-to-image and image-to-video replacement tasks. Our goal was to create a general pipeline for both data combinations and make the high-quality final result. As a baseline approach, we used the FaceShifter \cite{li2020faceshifter} model, which we updated with a new loss function and minor architecture enhancements. Furthermore, additional post-processing steps were developed and embedded in the pipeline to get result images or videos with high resolution.

\section{Related Work}
One of the first face swap approaches is based on a rather simple idea using autoencoders \cite{deepfakes}. In order to apply autoencoders to face swap the authors train two autoencoders: for the first person we want to transfer (source), and for the second one, the target itself. Both autoencoders have a common encoder, but different decoders. This model is trained in a simple way, with the exception of rare cases when we add some distortions to the input images in order to prevent the model from overfitting.

DeepFaceLab \cite{perov2021deepfacelab} is a well-known SOTA solution based on autoencoders implementation. In the paper the authors develop the idea mentioned above, and also add many other features for better transfer: additional training loss functions, slightly modified architecture, additional augmentations, and many other improvements. The main problem for this approach is that the model should be trained on source and target face images every time one wants to do face swap. So this approach can be rather applied for many real time tasks when we do not have immense datasets for training.

Another one approach that is able to process an image without retraining is the First Order Motion Model(FOMM) \cite{Siarohin_2019_NeurIPS}, as well as its development - Motion CoSegmentation \cite{Siarohin_2020_motion}. The idea of this method does not concern generating a new image from scratch, but in some sense to translate one frame into another using trainable segmentation maps and affine transformations. A dataset of a large number of videos was used as training data, and the training process was built on frames from one video, where the authors tried to translate one frame into another. However the idea looks promising the visual quality of synthesized results is far from perfect, because the model copes very poor with face rotations, so face recognition after transfer is inconvenient.

The following model \cite{zakharov2019fewshot} is based on the use of face keypoints for generation. Moreover this approach is the first in our survey that is based on Generative Adversarial Networks (GAN). The main idea of such networks is that we use two networks - a generator and a discriminator. The generator learns to generate realistic images, and the discriminator tries to distinguish the generated images from the real ones. The training process consists of two scenarios: firstly the generator tries to outperform the discriminator, and secondly the discriminator tries to detect synthesized images.

In the next article the authors \cite{li2020faceshifter} developed a model that shows some of the best results on face swap evaluation metrics. Two models are used in the proposed approach. The first model, the Adaptive Embedding Integration Network (AEI-Net), is used to perform the face transfer itself, and the second, the Heuristic Error Acknowledging Network (HEAR-Net), is used to improve the quality of the resulting transfer. We will describe AEI-Net in a bit more details further because this was our baseline.

In 2021 HifiFace model \cite{hififace2021} was proposed in order to produce a high quality face swap method. This model can preserve the face shape of the source face and generate photo-realistic results. The authors use 3D shape-aware identity to control the face shape instead of keypoint-based and feature-based methods for face areas. The methods show good results and preserve face identity with high quality.

The last but not least model that we would like to observe is the SimSwap model \cite{simswap2021}. Ideologically, the model is quite similar to FaceShifter \cite{li2020faceshifter}, and the difference is in using a common model architecture instead of two different models.

Although we can see very different approaches used to establish perfect visual quality of generated images, every method has its own pros and cons. In our research we tried to increase the quality of the generated images and at the same time to overcome some problems we found in recent articles and models. Further we provide a technical report of our solution and evaluation results to compare the proposed model with SOTA architectures. 

\section{The Proposed Approach}
\subsection{Baseline architecture}
Since we were faced with the task of building a model that would transfer only one image of a person, we immediately dismissed many existing solutions. FaceShifter \cite{li2020faceshifter}, SimSwap \cite{simswap2021} seemed to be the most relevant to the task being solved.

\begin{figure}[htbp]
\centerline{\includegraphics[width=0.5\textwidth]{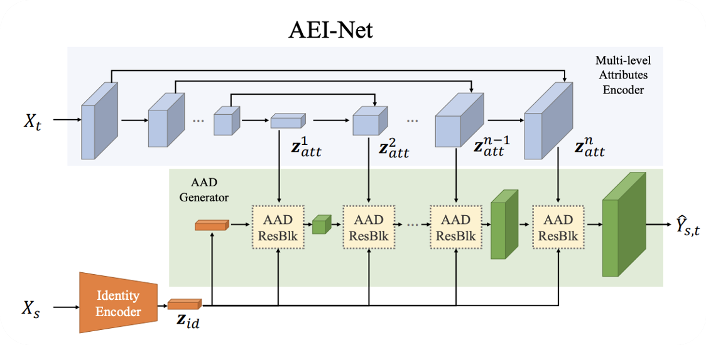}}
\caption{AEI-Net \cite{li2020faceshifter} architecture.}
\label{fig1}
\end{figure}

As a result, we took a part of the FaceShifter model - AEI-Net, but also implemented some new ideas to improve the final result. Just in case, we recall that AEI-Net works like this:
\begin{itemize}
    \item Identity (target) features extracted from $X_s$ using the pre-trained ArcFace model (Identity Encoder in Fig.~\ref{fig1})
    \item Attribute features extraction from $X_t$ - $z^{1}{att}, z^{2}{att}, ..., z^{n}_{att}$ using the U-Net-like architecture (blue part in Fig.~\ref{fig1})
    \item Attribute and identity features fusion
    \item Create output image $\hat{Y}_{s,t}$, which saves the 'identity' of a person from $X_s$ and attributes from $X_t$
\end{itemize}

At the top level, this whole architecture of our approach consists of three parts:
\begin{itemize}
    \item Identity encoder
    \item Encoder for $X_t$ - the blue part on the Fig.~\ref{fig1} that extracts attribute features from the target image
    \item A generator that generates the final image based on features from $X_t$ and identity vector
\end{itemize}

It should be mentioned that we also tried to evaluate different architectures for attribute extraction: U-Net, Linknet, Resnet. Overall U-Net provided the most promising results. Another experiment we conducted was on AAD blocks optimal number evaluation: as a result we used 2 AAD blocks in the released pipeline.

\subsection{Loss Function}

Choosing the right loss for the model is an essential step, because it tells us exactly what we want to achieve. In order to outperform the baseline approach we improved the loss function that was used with a number of additional features. This update provided our model with better performance in terms of quality. The list of baseline loss function parts is as follows:

\begin{itemize}
    \item $L_{id}$ represents identity loss. We assume that Identity Encoder outputs $\hat{Y}_{s,t}$ and $X_s$ values were close.
    \item $L_{adv}$ represents the GAN loss based on discriminator values (adversarial loss).
    \item $L_{rec}$ represents reconstruction loss. We use $X_s=X_t$ as model input randomly and require that the output value be $\hat{Y}_{s,t}=X_t$.
    \item $L_{att}$ represents attribute loss. We require that $z^{1}_{att}, z^{2}_{att}, ..., z^{n}_{att}$ values for $\hat{Y}_{s,t}$ and $X_t$ were close. 
\end{itemize}

Lets proceed to our loss modifications. First, we modified the reconstruction loss using the idea from the SimSwap \cite{simswap2021} architecture. In the original, the idea of this loss was that if we give the model two identical images of a person, we did not want the model to do something with the image. However, we went further here and did not require $X_s=X_t$, it was enough that $X_t$ and $X_s$ belong to the same person. In this case we required $X_t$ not to be changed in any way as a result of the transfer. Since we used datasets, where each person was presented with several frames, it became possible to implement such a modification of the loss.

Another one important modification was based on intuition that eyes appeared to be a very important component in the visual perception of the face swap output, especially when we use image-to-video transfer. In this case every single frame should represent the same line of sight for realistic perception. Therefore, we decided to add a special eye loss function which was obtained during experiments. It is based on L2 comparison of eyes areas features between ${X}_{t}$ and $\hat{Y}_{s,t}$, evaluated using face keypoints detection model.

\subsection{Image-to-Video Improvements}
When we perform face swap from image to video we save the transformation matrix for extracted faces on the frames. This information helps us insert the already changed face into its original place on the frame. However, if we insert the whole image obtained by our model, visual artifacts usually appear on the edge of the inserted area on the original frame and are clearly visible. This effect occurs both due to the incomplete correspondence of the brightness of the source image and target frame, and due to the possible blurring of the image synthesized by our model. Therefore, it is necessary to ensure a smooth transition from the source image to the resulting frame. For that reason we use segmentation masks.

\begin{figure}[htbp]
\centerline{\includegraphics[width=0.4\textwidth]{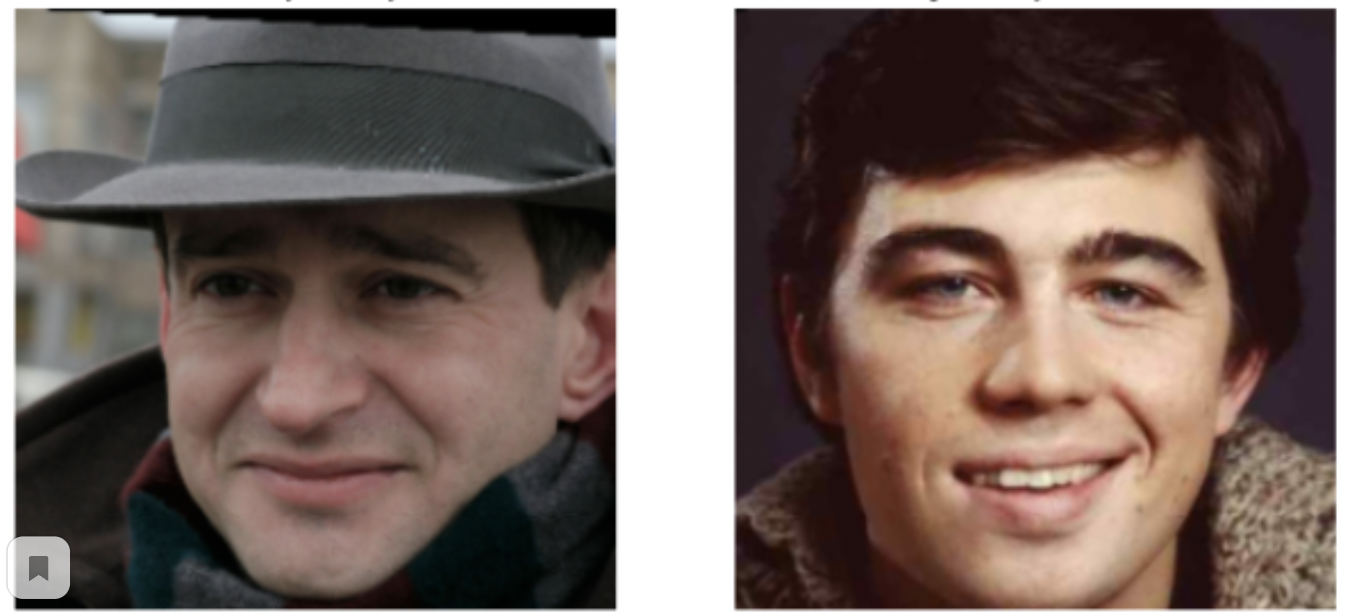}}
\centerline{\includegraphics[width=0.4\textwidth]{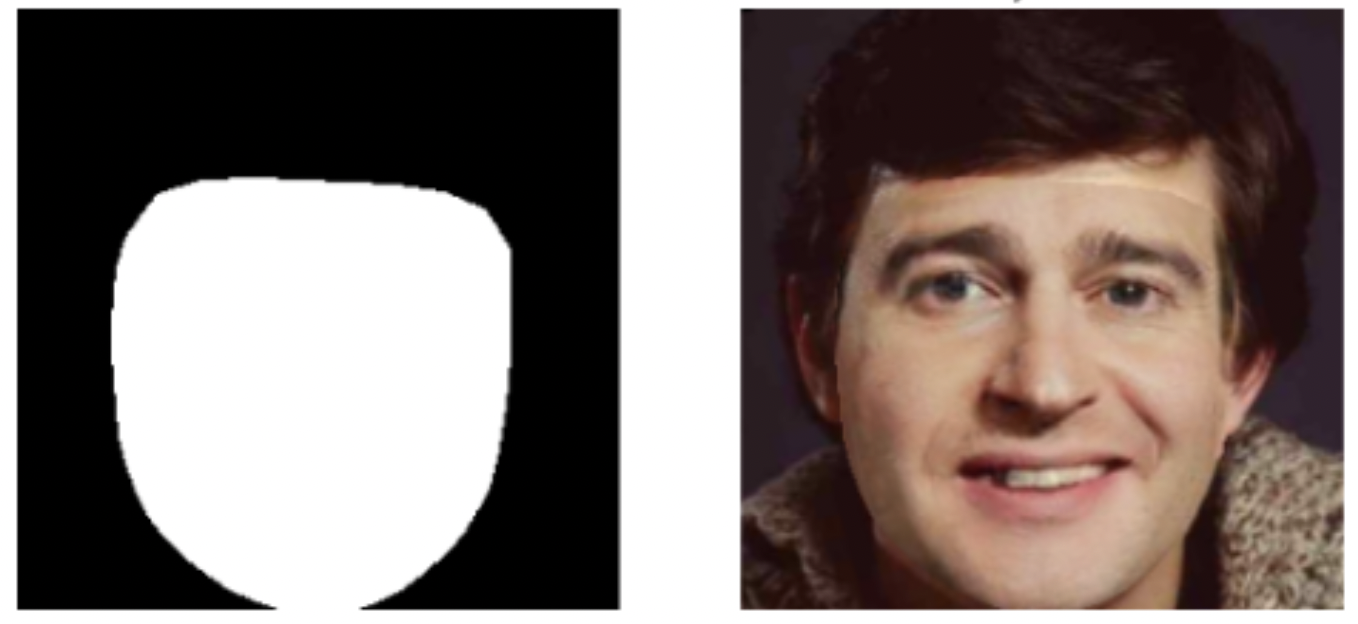}}
\centerline{\includegraphics[width=0.4\textwidth]{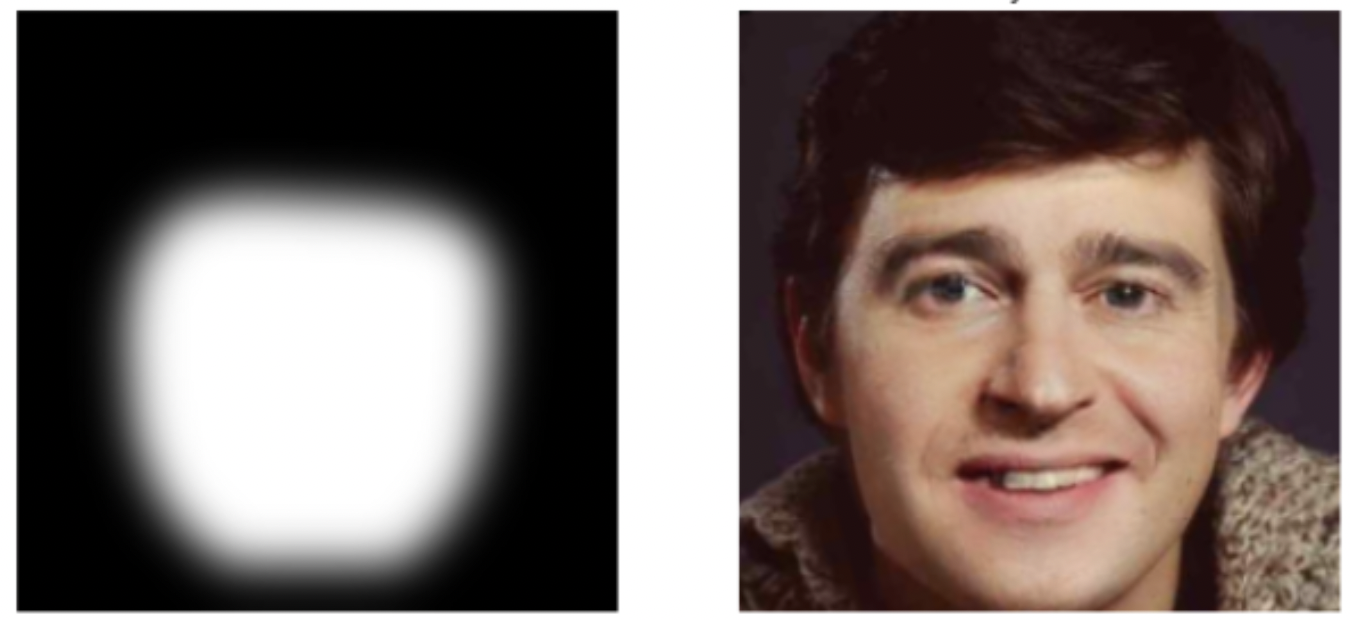}}
\caption{Face mask blurring effect in terms of face swap result.}
\label{fig2}
\end{figure}

A face mask is just a binary image that determines which pixels belong to the face and which do not. Thus, we can determine the exact location of the face and do precise contour crop. In order to lessen the effect of exact face area transfer we add Gaussian blurring at the edges. The result of such modification is presented in Fig.~\ref{fig2}. It can be also noted that the not only blurring added to the mask but also the mask area has changed. This is another modification we implemented to resolve the transfer problem for faces with distinct proportions. We explain it further.

At the experimental stage we encountered the following problem - sometimes $\hat{Y}_{s,t}$ and $X_t$ have distinct face proportions, as the model tries to keep the shape of the source face $X_s$. If the synthesized face $\hat{Y}_{s,t}$ is significantly wider than the target one $X_t$, then the transfer will be only partial, and we will not keep the shape of the source face $X_s$.

In order to deal with this problem we decided to track the keypoints for the generated face and the target face on the video. In case of the significant difference in the coordinates of the keypoints we modify the binary mask (compare the middle and bottom row masks in Fig.~\ref{fig2}). If the face obtained by the model completely covers the face in the video, we increase the mask, thereby creating the effect of transferring not only the face, but also the head's shape. Otherwise we reduce the mask and increase the blurring degree to transfer only the central part of the face.

\section{Evaluation}
\begin{figure}[htbp]
\centerline{\includegraphics[width=0.5\textwidth]{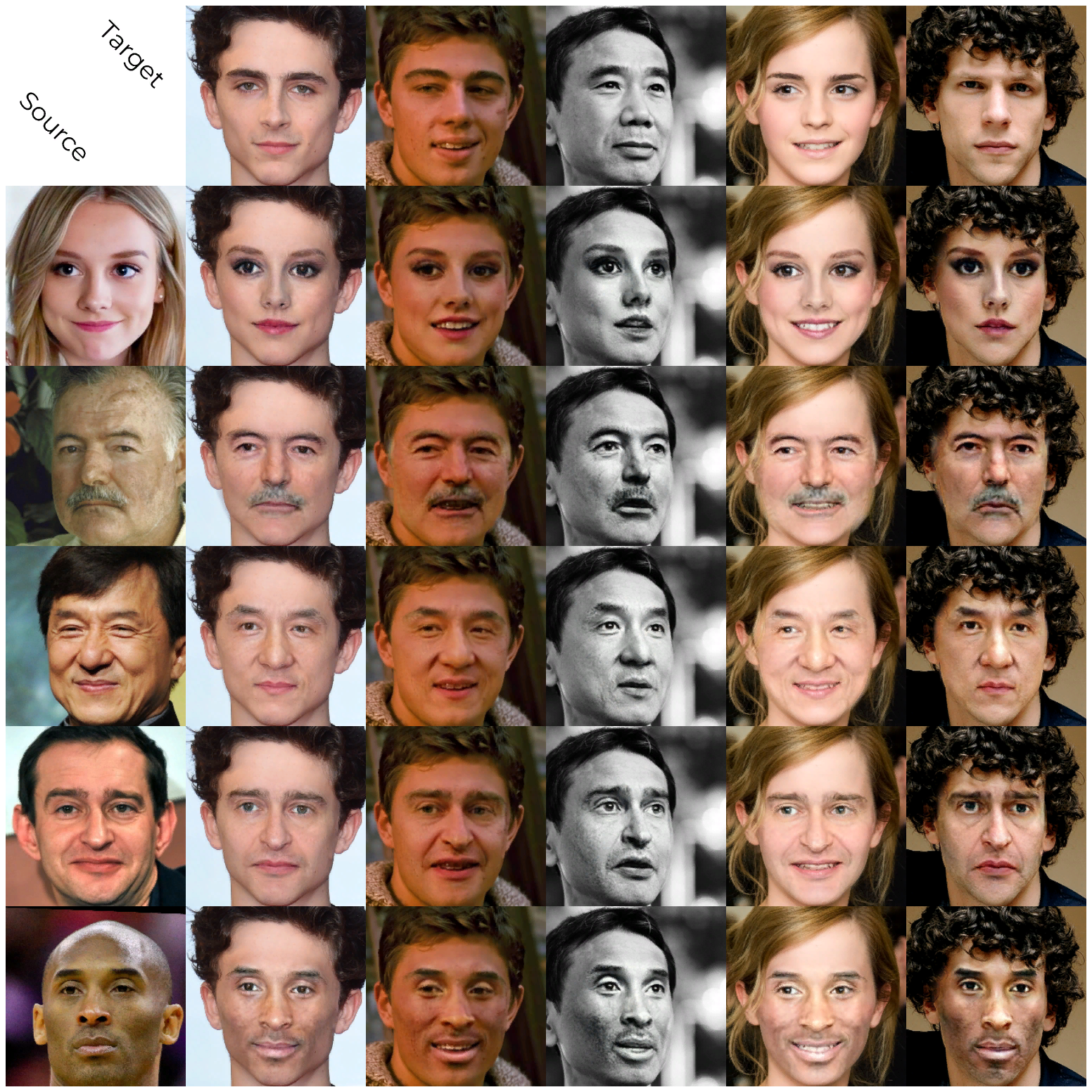}}
\caption{The proposed face swap model results.}
\label{fig3}
\end{figure}

In order to train and validate our model we selected two common datasets VGGFace2 \cite{vggface2} and CelebA-HQ \cite{liu2015faceattributes}. We used these datasets for training and further comparison of our model with SOTA architectures. VGGFace2 dataset meet our requirements due to ethnicity, gender, perspective and lightning conditions variability. We trained our model for 12 epochs with 19 batch size. Training experiments were carried out on the Tesla V100 32 GB GPU. Several frames were selected from the validation set to observe the quality of the proposed approach (Fig.~\ref{fig3}). Overall there were depicted 25 face swap results varying in face proportions, skin color, hair, etc. You can see how our model fits the source face to the target image.

Visual quality assessment is not the only way to estimate our face swap model results. We calculated several evaluation metrics to perform comparison with SOTA models. The metrics list was gathered from FaceShifter, SimSwap and HifiFace articles. Without going into mathematical details here we present a list of metrics used for evaluation:

\begin{itemize}
    \item ID\_retrieval and shape\_ringnet - responsible for preserving identity (head shape, etc.)
    \item Exp\_ringnet - responsible for facial expression and saving emotions.
    \item Eye\_ldmk - for maintaining the direction of view.
\end{itemize}

All the experiments we conducted allowed us to conclude that the proposed model is well-trained and can be used in different cases like image-to-image and image-to-video transfer with high quality. In the following Table~\ref{tab1} we can observe a comparison of our model with other approaches in terms of identity and attributes encoders (before blending). Here we use our conventional model with a U-Net encoder and two AAD blocks.

\begin{table}[htbp]
\caption{The proposed approach comparison with SOTA models (before blending)}
\begin{center}
\begin{tabular}{|c|c|c|}
\hline
\textbf{Method} & \textbf{Identity retrieval} & \textbf{Attributes retrieval} \\
\hline
FaceSwap (2017) & 54.19 & 2.51 \\
DeepFakes (2018) & 77.65 & 4.59 \\
FaceShifter (2019) & 97.38 & 2.96 \\
SimSwap (2021) & 92.83 & 1.53 \\
HifiFace (2021)  & 98.48 & 2.63 \\
Ours & 98.67 & 3.00\\
\hline
\end{tabular}
\label{tab1}
\end{center}
\end{table}

\begin{table*}[t]
\caption{The proposed approach comparison with SOTA models (after blending)}
\begin{center}
\begin{tabular}{|c|c|c|c|c|c|c|}
\hline
\textbf{Method} & \textbf{Identity retrieval} & \textbf{Shape\_ringnet} & \textbf{Exp\_ringnet} & \textbf{Pose\_ringnet} & \textbf{Shape\_HN} & \textbf{Eye\_ldmk} \\
\hline

FaceSwap (2017)	& 58.82 & 0.75 & 0.305 & 0.045 & 1.94 & 4.22 \\
DeepFakes (2018) & 72.42 & 0.65 & 0.696 & 0.110 & 7.35 & 11.6 \\
FaceShifter (2019) & 82.02 & 0.67 & 0.420 & 0.040 & 1.93 & 2.48 \\
SimSwap (2021) & 87.42 & 0.72 & 0.340 & 0.035 & 2.13 & 2.91 \\
HifiFace (2021) & 89.17 & 0.64 & 0.510 & 0.048 & 2.13 & 2.04 \\
Ours & 90.61 & 0.64 & 0.436 & 0.047 & 2.26 & 2.02 \\

\hline
\end{tabular}
\label{tab2}
\end{center}
\end{table*}

We also calculated the evaluation metrics independently for each method after blending. In order to compare all the models in a consistent manner we used the provided videos in the FaceForensics++ dataset \cite{faceforensics}. The results are provided in Table~\ref{tab2}.

\section{Conclusion}
In order to make conclusion we should mention that our model outperforms many SOTA architectures in terms of several well-known metrics. At the same time the visual quality of the generated results also proves that fact. Several new features made the proposed pipeline suitable for image-to-image face swap as well as image-to-video: general architecture based on AEI-Net, new eye loss, super resolution and face mask tuning based on source/target faces area proportion analysis. On the contrary, many SOTA architectures are evaluated only in image domain and are not suitable for videos processing. We also shared the trained model publicly on \href{https://github.com/sberbank-ai/sber-swap}{GitHub} and \href{https://colab.research.google.com/drive/1B-2JoRxZZwrY2eK_E7TB5VYcae3EjQ1f}{Google Colab}.

\printbibliography

\end{document}